\documentclass[10pt,twocolumn,letterpaper]{article}

\usepackage{cvpr}              

\usepackage{makecell}
\usepackage{bbm}

\definecolor{cvprblue}{rgb}{0.21,0.49,0.74}
\usepackage[pagebackref,breaklinks,colorlinks,allcolors=cvprblue]{hyperref}

\def\paperID{18142} 
\def\confName{CVPR}
\def\confYear{2025}

\title{Boosting MLLM Reasoning with Text-Debiased Hint-GRPO}

\author{
{Qihan Huang}\textsuperscript{1}, {Weilong Dai}\textsuperscript{2},
{Jinlong Liu}\textsuperscript{2}, {Wanggui He}\textsuperscript{2}, {Hao Jiang}\textsuperscript{2},\\
{Mingli Song}\textsuperscript{1}, {Jingyuan Chen}\textsuperscript{1}, {Chang Yao}\textsuperscript{1}, {Jie Song}\textsuperscript{1}\\
    \textsuperscript{1} Zhejiang University,
    \textsuperscript{2} Alibaba Group\\
    {\tt\small \{qh.huang,brooksong,jingyuanchen,changy,sjie\}@zju.edu.cn,} \\
    {\tt\small \{chenlong0104.chen,aoshu.jh\}@alibaba-inc.com,} \\
    {\tt\small LJLwykqh@126.com,wanggui.hwg@taobao.com}
}

\begin{document}
\maketitle


\begin{abstract}
MLLM reasoning has drawn widespread research for its excellent problem-solving capability.
Current reasoning methods fall into two types: PRM, which supervises the intermediate reasoning steps, and ORM, which supervises the final results.
Recently, DeepSeek-R1 has challenged the traditional view that PRM outperforms ORM, which demonstrates strong generalization performance using an ORM method~(\ie, GRPO).
However, current MLLM's GRPO algorithms still struggle to handle challenging and complex multimodal reasoning tasks~(\eg, mathematical reasoning).
In this work, we reveal two problems that impede the performance of GRPO on the MLLM: Low data utilization and Text-bias.
Low data utilization refers to that GRPO cannot acquire positive rewards to update the MLLM on difficult samples, and text-bias is a phenomenon that the MLLM bypasses image condition and solely relies on text condition for generation after GRPO training.
To tackle these problems, this work proposes Hint-GRPO that improves data utilization by adaptively providing hints for samples of varying difficulty, and text-bias calibration that mitigates text-bias by calibrating the token prediction logits with image condition in test-time.
Experiment results on three base MLLMs across eleven datasets demonstrate that our proposed methods advance the reasoning capability of original MLLM by a large margin, exhibiting superior performance to existing MLLM reasoning methods.
Our code is available at \textit{~\url{https://github.com/hqhQAQ/Hint-GRPO}}.

\end{abstract}

\section{Introduction \label{sec:intro}}

MLLM~(Multimodal LLM) reasoning has attracted wide research interest for its exceptional problem-solving capability, especially after the release of OpenAI's o1 model.
Existing reasoning methods can be categorized into two types: PRM~(Process Reward Method), which supervises the intermediate reasoning steps, and ORM~(Outcome Reward Method), which supervises the final reasoning results.
Previously, most methods~\cite{xu2024llava, yao2024mulberry, liu2024diving, dong2024progressive} consider PRM to be superior to ORM, leveraging numerous strategies~(\eg, MCTS~\cite{yao2024mulberry, liu2024diving, dong2024progressive}, DPO~\cite{zhang2024improve}) for PRM training.

\begin{figure}[t]
\centering
    \includegraphics[width=1.0\linewidth]{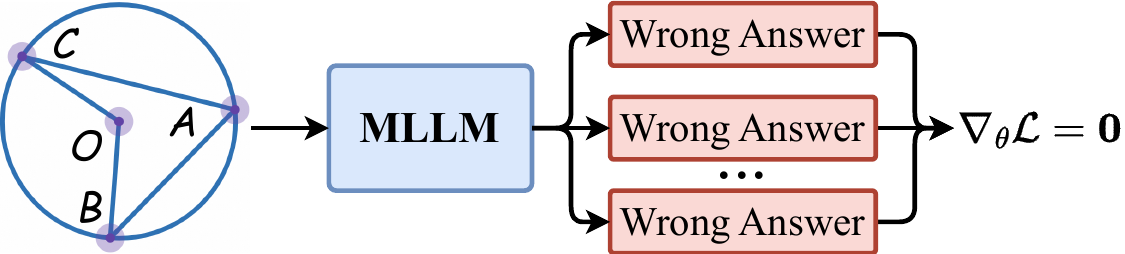}
\caption{{\textbf{Low data utilization} of GRPO: If all answers are incorrect, the zero loss gradients~($\nabla_\theta\mathcal{L}=\mathbf{0}$) will invalidate the sample.}}
\label{fig:intro_invalid_sample}
\end{figure}
\begin{figure}[t]
\centering
    \includegraphics[width=1.0\linewidth]{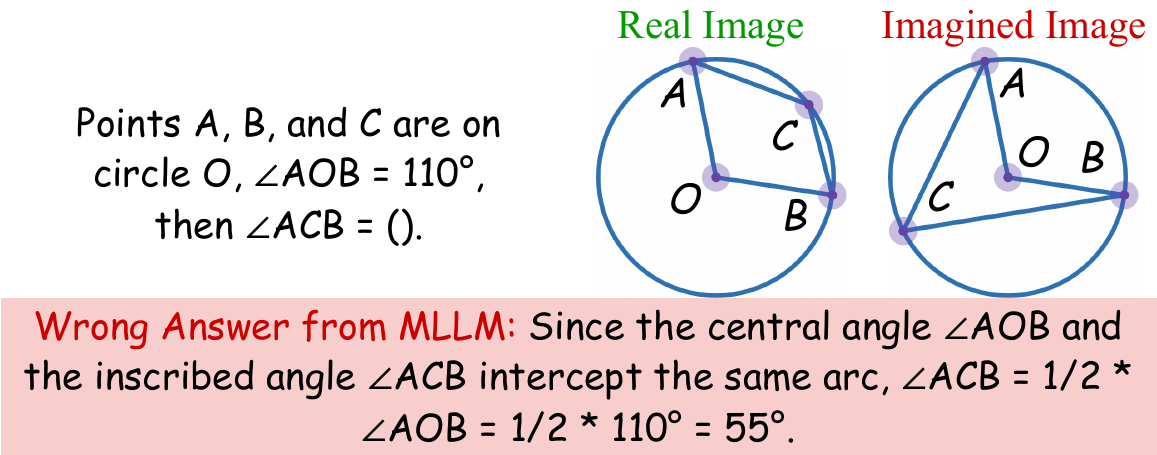}
\caption{\textbf{Text-bias} of GRPO, where the MLLM ignores real image and uses its imagined image from text to generate outputs.}
\vspace{-1em}
\label{fig:intro_text_bias}
\end{figure}

Recently, DeepSeek-R1~\cite{guo2025deepseek} has overturned the conventional belief that PRM is superior to ORM, which employs the GRPO~(Group Relative Policy Optimization~\cite{shao2024deepseekmath}) reinforcement learning algorithm that calculates rewards for model predictions solely based on the model's reasoning results, without supervision of intermediate reasoning steps.
DeepSeek-R1 discovers that ORM training can induce models to engage in self-reflection, extending the chain of thought to reach correct answers, thereby achieving excellent reasoning performance.
After the emergence of DeepSeek-R1, many researchers explore ORM methods for LLM \& MLLM reasoning, with most focusing on the GRPO algorithm~(\eg, Open-R1~\cite{openr1}, simpleRL-reason~\cite{zeng2025simplerl}, R1-V~\cite{chen2025r1v}, R1-Multimodal-Journey~\cite{r1-multimodal-journey}).
Some research~(Open-Reasoner-Zero~\cite{OpenReasonerZero2025}) also finds that PPO-based~\cite{schulman2017proximal} ORM algorithm could achieve similar performance to GRPO.

However, while current MLLM's ORM methods~(\eg, R1-V~\cite{chen2025r1v}) perform well on simple visual tasks~(\eg, counting objects in the image), they are insufficient for more challenging and complex multimodal reasoning tasks~(\eg, mathematical reasoning).
In this work, we identify two problems that hinder the performance of ORM methods on MLLM reasoning: \textbf{(1) Low data utilization}; \textbf{(2) Text-bias}.

\begin{figure}[t]
\centering
    \includegraphics[width=1.0\linewidth]{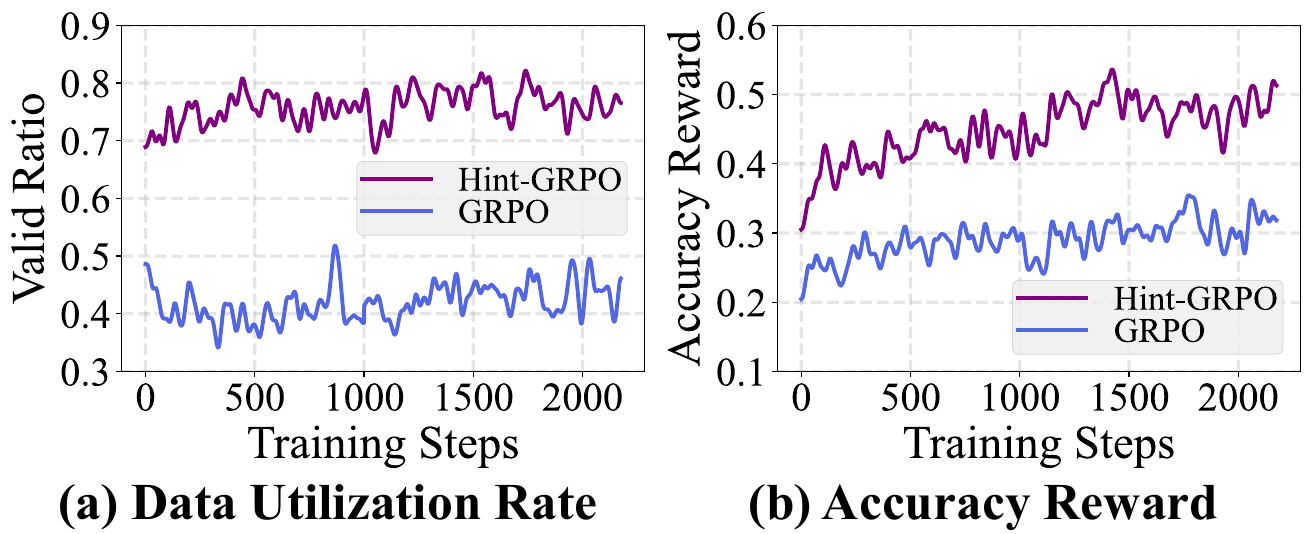}
\caption{Data utilization rate \& reward of GRPO \& Hint-GRPO.}
\label{fig:intro_ratio_reward}
\end{figure}

For the first problem, GRPO requires to first prompt the MLLM to generate multiple predictions for the same question, and then assign rewards to the predictions based on their correctness.
However, due to the insufficient reasoning capability of the original MLLM, all the predictions for the difficult question could be incorrect in MLLM.
In this situation, GRPO cannot update the model as the calculated advantages in GRPO are all zero, rendering the training sample invalid, as shown in \autoref{fig:intro_invalid_sample}.
\autoref{fig:intro_ratio_reward}~(a) also shows the ratio of valid samples in each batch during training~(Qwen2-VL-7B~\cite{wang2024qwen2} on the mathematical reasoning dataset), verifying the severity of low data utilization.

To address this problem, this work proposes \textit{Hint-GRPO}, which provides additional hints for MLLM to solve questions of high difficulty.
Specifically, for difficult questions where MLLM cannot find the correct answer~(invalid samples), Hint-GRPO provides the initial part of correct reasoning steps~(according to a certain ratio) to the MLLM as hints, allowing it to complete the remaining reasoning steps and arrive at the final answer.
As shown in \autoref{fig:intro_ratio_reward}~(a), Hint-GRPO effectively leads MLLM to generate correct answers, thereby improving the data utilization.
Furthermore, Hint-GRPO can adaptively adjust hint ratios for questions of varying difficulty, \textbf{avoiding excessive hints} for simple questions, thereby allowing for optimal utilization of the dataset and achieving better performance.

For the second problem~(text-bias), this work observes a phenomenon that during GRPO training, MLLM learns to directly infer the final answer from text condition while ignoring image condition, as shown in \autoref{fig:intro_text_bias}.
Besides, \autoref{fig:acc_ratio}~(a) demonstrates that as GRPO training progresses, the accuracy of MLLM~(with the image condition removed) on the test set also increases.

To tackle this problem, this work proposes a \textit{text-bias calibration} method \textbf{in test-time} to reduce MLLM reasoning errors caused by ignoring image condition.
Inspired by CFG~(classifier-free guidance)~\cite{ho2022classifier} in image generation, this text-bias calibration method first uses the MLLM to generate prediction results~(token logits) with and without image condition, and then calibrates the token logits using the difference between them.

We perform comprehensive experiments to validate the performance of the proposed methods.
Specifically, we apply text-debiased Hint-GRPO to three base MLLMs across eleven datasets~(mathematical reasoning \& universal multimodal reasoning), and the experiment results demonstrate that our methods improve the reasoning capability of original MLLM by a large margin, achieving significantly superior performance to existing MLLM reasoning methods.

To sum up, the main contributions of this work can be summarized as follows:

$\bullet$ We identify and thoroughly analyze two problems that hinder the performance of GRPO on MLLM reasoning: (1) Low data utilization; (2) Text-bias.

$\bullet$ We propose two methods~(Hint-GRPO and text-bias calibration), which effectively mitigate these two problems.

$\bullet$ Experiment results show that our proposed methods achieve significantly superior performance to existing MLLM reasoning methods.

\begin{figure*}[t]
\centering
    \includegraphics[width=1.0\linewidth]{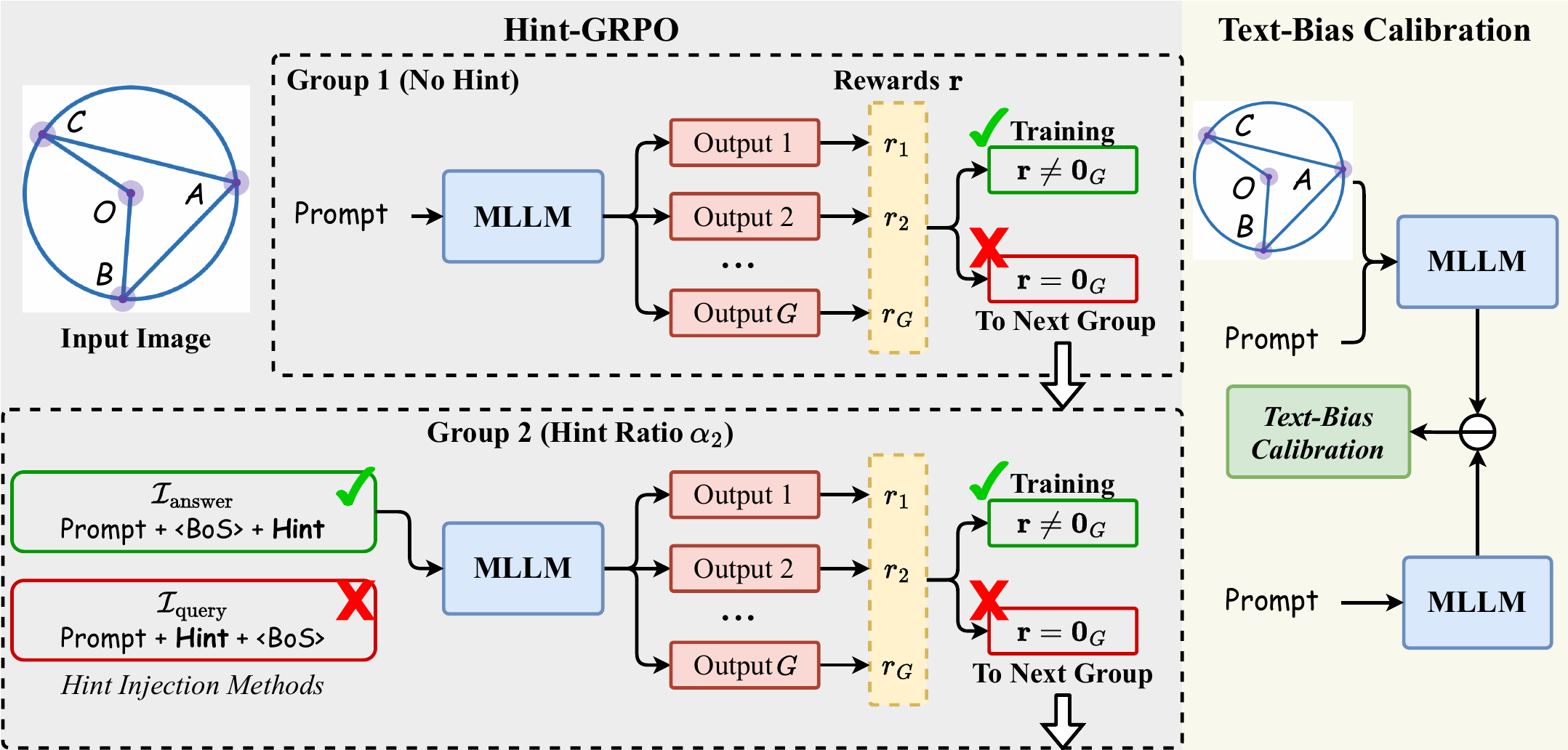}
\caption{Framework of Hint-GRPO and text-bias calibration. Specifically, Hint-GRPO adaptively provides hints to the samples and selects the most eligible group for training, mitigating the low data utilization problem. Text-bias calibration increases the intensity of image condition \textbf{in test-time}, alleviating the text-bias problem. Note that ``$<$BoS$>$'' denotes the \textit{beginning-of-sequence} token in MLLM.
}
\vspace{-1em}
\label{fig:framework}
\end{figure*}

\section{Related Work}

\noindent \textbf{PRM for MLLM Reasoning.}
Most existing MLLM reasoning methods follow the PRM~(Process Reward Method) paradigm, which employs fine-grained supervision on the intermediate reasoning steps.
LLaVA-o1~\cite{xu2024llava} trains the MLLM to generate structured reasoning steps.
AtomThink~\cite{xiang2024atomthink} directly trains a process reward model to assess the reasoning steps, using the framework of Markov decision process~(MDP) learning.
Mulberry~\cite{yao2024mulberry}, M-STAR~\cite{liu2024diving}, and AR-MCTS~\cite{dong2024progressive} utilize Monte Carlo Tree Search~(MCTS) to estimate the reasoning steps.
LLaVA-Reasoner-DPO~\cite{zhang2024improve} employs direct preference optimization~(DPO)~\cite{rafailov2023direct} to improve the intermediate reasoning process.
Virgo~\cite{du2025virgo} constructs a long-thought dataset to enable the MLLM with long reasoning capability.
LlamaV-o1~\cite{thawakar2025llamav} adopts curriculum learning to train the MLLM in an easy-to-hard manner.
However, DeepSeek-R1~\cite{guo2025deepseek} points out current PRM methods struggle to accurately evaluate the reasoning steps and suffer from a serious reward hacking problem, resulting in suboptimal performance.
On the contrary, ORM enables accurate evaluation by simply comparing the reasoning results with the ground truth, thus avoiding the reward hacking problem.

\vspace{0.5em}
\noindent \textbf{ORM for MLLM Reasoning.}
ORM~(Outcome Reward Method) only supervises the reasoning results using the ground truth, regardless of the intermediate reasoning steps.
ORM methods such as GRPO can achieve strong generalization performance, verified by DeepSeek-R1.
After the emergence of DeepSeek-R1, more researchers are shifting towards the ORM paradigm~(\ie, GRPO) for MLLM reasoning.
Specifically, Open-R1-Multimodal~\cite{open-r1-multimodal} establishes the first GRPO baseline for MLLM reasoning.
R1-V~\cite{chen2025r1v} demonstrates that GRPO performs well in simple multimodal reasoning tasks, \eg, counting objects in the image.
R1-Multimodal-Journey~\cite{r1-multimodal-journey} significantly accelerates the training speed using the vLLM package~\cite{kwon2023efficient}.
Video-R1~\cite{Video-R1} generalizes Open-R1-Multimodal to video reasoning.
However, current methods are still limited in addressing more challenging multimodal reasoning tasks~(\eg, mathematical reasoning).
Our work identifies two problems that hinder the GRPO performance in MLLM reasoning~(low data utilization and text-bias), and proposes two methods to address them~(Hint-GRPO and text-bias calibration).

\section{Method \label{sec:method}}


\subsection{Preliminaries \label{sec:preliminaries}}

\noindent \textbf{Supervised Fine-tuning~(SFT).}
SFT trains the LLM on curated query-output pairs to improve its instruction-following ability.
The objective of SFT is to maximize the following objective:

\vspace{-0.5em}
\begin{equation}
\label{equa:sft_loss}
    \mathcal{J}_{\rm SFT}(\theta) \!=\! \mathbb{E}[q, o \! \sim \! P(Q,O)] \! \left( \! \frac{1}{|o|} \sum_{t=1}^{|o|} \log \pi_\theta(o_t|q, o_{<t}) \! \right) \!,
\nonumber
\end{equation}

where $q, o$ is the query-output pair sampled from the SFT dataset $P(Q,O)$, $\theta$ denotes the model parameters, $\pi_\theta(o_t|q, o_{<t})$ represents the logit of the model predicting the next token $o_t$ from $q$ and previous tokens $o_{<t}$.

\vspace{0.5em}
\noindent \textbf{Proximal Policy Optimization~(PPO).}
PPO is an actor-critic RL algorithm that is widely used in the RL fine-tuning stage of LLM.
In particular, it optimizes the model by maximizing the following objective:

\vspace{-0.5em}
{\small
\begin{equation}
\label{equa:ppo_loss}
    \mathcal{J}_{\rm PPO}(\theta) \! = \! \mathbb{E}[q \! \sim \! P(Q), \! o \! \sim \! \pi_{\theta_{\rm old}}(O|q)] \frac{1}{|o|} \! \sum_{t=1}^{|o|} \! \frac{\pi_\theta(o_t|q, o_{<t})}{\pi_{\theta_{\rm old}}(o_t|q, o_{<t})} A_t,
\nonumber
\end{equation}
}

where $\pi_{\theta_{\rm old}}$ is the old model, $A_t$ represents the advantage function measuring how the $t$-th token's prediction deviates from average, based on the rewards $\{ r_{\geq t} \}$ and a learned value function $V_{\psi}$.
The \textbf{min \& clip operations} for avoiding extreme values are omitted here for simplicity.

\vspace{0.5em}
\noindent \textbf{Group Relative Policy Optimization~(GRPO).}
GRPO directly calculates $A_t$ using the average reward of multiple sampled outputs, eliminating the additional value function $V_{\psi}$ in PPO.
Specifically, GRPO samples a group of outputs $\{ o_1, o_2, ..., o_G \}$ from the old model $\pi_{\theta_{\rm old}}$, and then optimizes the model by maximizing the following objective~(\textbf{min \& clip operations} are also omitted here):

\vspace{-0.5em}
{\small
\begin{align}
& \mathcal{J}_{\rm GRPO}(\theta) = \mathbb{E}[q \sim P(Q), \{o_i\}_{i=1}^G \sim \pi_{\theta_{\rm old}}(O|q)] \nonumber \\
& \frac{1}{G} \sum_{i=1}^G \frac{1}{|o_i|} \sum_{t=1}^{|o_i|} \left\{ \frac{\pi_\theta(o_{i,t}|q,o_{i,<t})}{\pi_{\theta_{\rm old}}(o_{i,t}|q,o_{i,<t})} \hat{A}_{i,t} - \beta D_{\rm KL}[\pi_\theta||\pi_{\rm ref}] \right\}, \nonumber
\end{align}
}

where $D_{\rm KL}[\pi_\theta||\pi_{\rm ref}]$ serves as a regularization term that prevents the new model $\pi_\theta$ from deviating too far from the original model $\pi_{\rm ref}$~(the model before training).
As an ORM method, GRPO provides the reward $r_i$ at the end of each output $o_i$~($r_i = 1$ if the reasoning result is correct, otherwise $r_i = 0$), and sets the advantage $\hat{A}_{i,t}$ of all tokens in $o_i$ as the normalized reward~($\mathbf{r} = \{ r_1, r_2, ..., r_G \}$, $\text{mean}(\cdot)$ denotes the average, and $\text{std}(\cdot)$ denotes the standard deviation):

\vspace{-0.5em}
\begin{equation}
\label{equa:advantage}
    \hat{A}_{i,t} = \tilde{r}_i = \frac{r_i - \text{mean}(\mathbf{r})}{\text{std}(\mathbf{r})}.
\end{equation}

\subsection{Hint-GRPO}


\subsubsection{Low Data Utilization}

Low data utilization refers to the problem that in current GRPO training, many training samples fail to provide effective feedback to the MLLM, as shown in \autoref{fig:intro_invalid_sample}.
Specifically, due to the limited reasoning ability of the original MLLM, the generated reasoning results on these training samples are \textbf{all incorrect}.
This situation~(\ie, $r_i = 0$ for all $\{ r_i \}_{i=1}^{G}$) causes each advantage $\hat{A}_{i,t} = \frac{r_i - \text{mean}(\mathbf{r})}{\text{std}(\mathbf{r})} = 0$, thus hindering the MLLM training~(note that $\text{std}(\mathbf{r})$ has a small offset to avoid division by zero).
Furthermore, we can analyze the invalidity of these training samples by examining the gradients of the optimizing objective $\mathcal{J}_{\rm GRPO}(\theta)$ on the MLLM parameters $\theta$:

\vspace{-0.5em}
{\small
\begin{align}
& \nabla_\theta \mathcal{J}_{\rm GRPO}(\theta) = \mathbb{E}[q \sim P(Q), \{o_i\}_{i=1}^G \sim \pi_{\theta_{\rm old}}(O|q)] \nonumber \\
& \frac{1}{G} \! \sum_{i=1}^G \! \frac{1}{|o_i|} \! \sum_{t=1}^{|o_i|} \! \left\{ \! \frac{\hat{A}_{i,t} \nabla_\theta \pi_\theta(o_{i,t}|q,o_{i,<t})}{\pi_{\theta_{\rm old}}(o_{i,t}|q,o_{i,<t})} \! - \! \beta \nabla_\theta D_{\rm KL}[\pi_\theta||\pi_{\rm ref}] \! \right\}. \nonumber
\end{align}
}

When all $\hat{A}_{i,t}$ equal 0, the update of MLLM parameters $\theta$ only depends on the less important KL divergence $D_{\rm KL}[\pi_\theta||\pi_{\rm ref}]$ and is unrelated to the accuracy of reasoning results.
This prevents such training samples from providing effective feedback for model optimization.

In this work, we conduct an in-depth analysis of the low data utilization problem.
Specifically, we propose a \textit{data utilization rate} to measure the proportion of effective samples in each batch during GRPO training.
Let $\{ z_k \}_{k=1}^{B}$ denote a training batch of $B$ samples, and the MLLM generates $G$ outputs for each sample, then $\mathbf{r}(z_k) \in \mathbb{R}^{G}$ represents the correctness of all $G$ outputs of $z_k$, \ie, $\mathbf{r}(z_k)_i = 1$ if the output result is correct, otherwise $\mathbf{r}(z_k)_i = 0$.
Next, we can determine the sample $z_k$ is valid \textbf{if and only if} $\text{std}(\mathbf{r}(z_k)) \ne 0$.
In other words, $\text{std}(\mathbf{r}(z_k)) = 0$ indicates that $\mathbf{r}(z_k)$ equals a zero vector $\mathbf{0}_G$ or a ones vector $\mathbf{1}_G$, resulting in the calculated advantages $\hat{A}_{i,t}$ being all zero and thus invalidating the sample $z_k$.
Finally, the data utilization rate $S_{\rm valid} \in [0, 1]$ for this training batch is calculated as below~($\mathbbm{1}\{ \cdot \}$ denotes the indicator function):

\vspace{-0.5em}
\begin{equation}
\label{equa:data_utilization_rate}
    S_{\rm valid} = \frac{1}{B} \sum\limits_{k=1}^{B} \mathbbm{1} \{ \text{std}(\mathbf{r}(z_k)) \ne 0 \}.
\end{equation}

\begin{figure}[t]
\centering
    \includegraphics[width=1.0\linewidth]{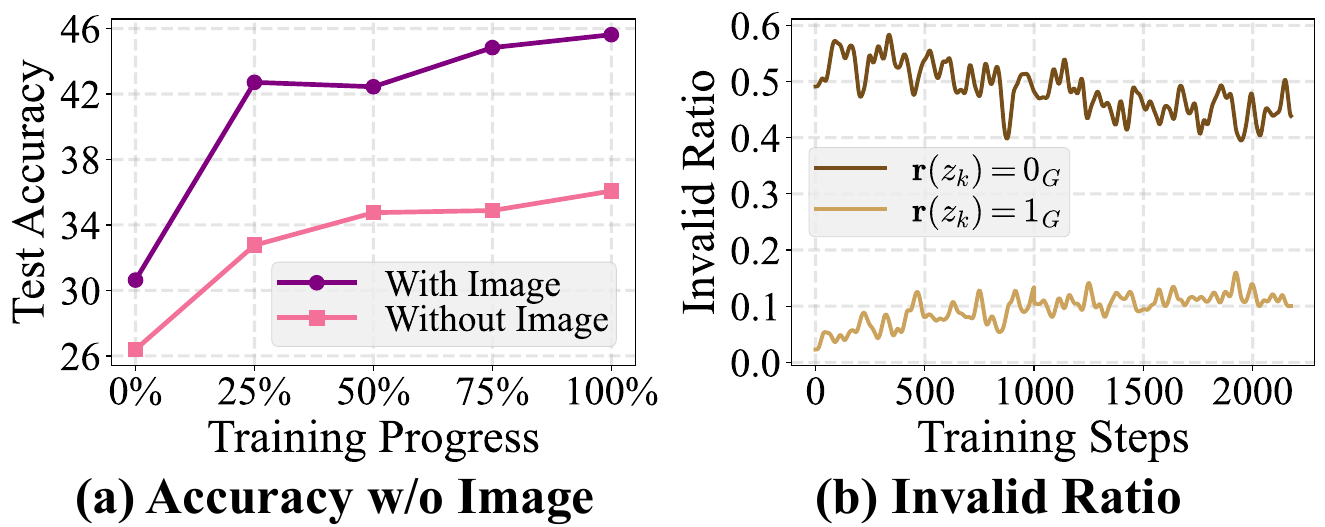}
\caption{(a) Qwen2-VL-7B's test accuracy w/ \& w/o image in GRPO training. (b) The ratio of two types of invalid samples.}
\label{fig:acc_ratio}
\end{figure}

As shown in \autoref{fig:intro_ratio_reward}~(a), the original GRPO exhibits a low $S_{\rm valid}$~(40\% to 50\%) for Qwen2-VL-7B during training.
In addition, we also calculate the ratio of two situations where the training sample is invalid:
(1) $\mathbf{r}(z_k)$ equals $\mathbf{0}_G$, \ie, all outputs of $z_k$ are incorrect.
(2) $\mathbf{r}(z_k)$ equals $\mathbf{1}_G$, \ie, all outputs of $z_k$ are correct.
\autoref{fig:acc_ratio}~(b) presents that situation~(2) only accounts for a small proportion~(within 10\%), which corresponds to the phenomenon of low rewards during GRPO training in \autoref{fig:intro_ratio_reward}~(b).
These findings verify that GRPO suffers from the low correctness of MLLM outputs, thus constraining the training of MLLM reasoning.

\subsubsection{Hint-GRPO Implementation}

To tackle the problem of low data utilization, this work proposes Hint-GRPO, which provides \textbf{reasoning hints} to lead the MLLM to generate correct answers for difficult training samples, thus improving the number of samples with $\text{std}(\mathbf{r}(z_k)) \ne 0$.
Specifically, let $h$ denote the reasoning hint, then the MLLM predicts $\pi_\theta(o_{i,t}|q,h,o_{i,<t})$~(the next-token prediction logit) with $h$ as condition in Hint-GRPO.
The implementation of Hint-GRPO consists of three parts: (1) Dataset construction; (2) Hint injection method; and (3) Hint adaptation strategy.

\vspace{0.5em}
\noindent \textbf{(1) Dataset construction}

\noindent The training of Hint-GRPO requires a dataset containing multiple samples with image, query text, reasoning steps, and the ground-truth answer.
Here, we choose the LLaVA-CoT dataset~\cite{xu2024llava}, a high-quality and influential training dataset for MLLM reasoning, as our base dataset.
Nevertheless, the original LLaVA-CoT dataset still necessitates \textbf{two modifications} before being used for Hint-GRPO:

First, we use GPT-4o to split the original reasoning steps~(a long text) of each sample into multiple structured steps, which allows us to control the hint level by directly adjusting the number of reasoning steps in the hint.

Second, we convert multi-choice questions to fill-in-the-blank format by removing the options, stopping the MLLM from getting correct answers through random guessing.

\vspace{0.5em}
\noindent \textbf{(2) Hint injection method}

\noindent In $\pi_\theta(o_{i,t}|q,h,o_{i,<t})$~(the next-token prediction logit of Hint-GRPO), the MLLM requires to inject $h$~(the reasoning hint split from the correct reasoning steps in the dataset) into the model.
The simplest baseline $\mathcal{I}_{\rm query}$~(named \textit{hint injection in query}) in \autoref{fig:framework} is to append the hint to the original query text, \eg, appending ``Solve the question following the hint: \{\textit{Reasoning Hint}\}''.
However, this baseline has two problems:
First, even with hints in the prompt, the model sometimes ignores them and starts reasoning from scratch.
Second, and more critically, query text \textbf{with hint} in training-time is inconsistent with query text \textbf{without hint} in test-time, leading to poor test performance.

To address these problems, we propose $\mathcal{I}_{\rm answer}$~(named \textit{hint injection in answer}) in \autoref{fig:framework}, which keeps the query text unchanged while using the reasoning hint as the beginning of model output, letting the model complete the remaining steps to reach the final answer.
$\mathcal{I}_{\rm answer}$ makes the MLLM faithfully continue reasoning based on the hint, and the query text requires no hint in both training and testing, performing significantly better than the baseline $\mathcal{I}_{\rm query}$.

\vspace{0.5em}
\noindent \textbf{(3) Hint adaptation strategy}

\noindent Hint-GRPO requires adjusting the hint level to achieve optimal results. Specifically, if the hint level is too low, it still cannot lead the MLLM to the correct answers, remaining trapped in the low data utilization problem;
If the hint level is too high, the MLLM can reach correct answers without reasoning, hindering effective use of the data for MLLM reasoning training.
Therefore, based on our constructed dataset, we employ a hint ratio $\alpha \in [0,1]$ to adjust the hint level.
In detail, for the total $L$ correct reasoning steps of a sample, we extract the \textbf{first} $L \cdot \alpha$ steps as hint~(denoted as $h_\alpha$), and thus a higher $\alpha$ indicates a higher hint level.
Next, we explore three strategies to set the hint level: (1) Fixed hint level; (2) Random hint level; (3) Adaptive hint level.

\textbf{Fixed hint level} sets the same hint ratio $\alpha$ for each training sample.
However, this strategy can lead to simple questions being given excessive hints, resulting in insufficient training for MLLM reasoning.
Experiments in \autoref{tab:ablation_dataset_injection} show that this strategy achieves suboptimal performance.    

\textbf{Random hint level} samples $\alpha \sim \text{Uniform}(0, 1)$ randomly for each training sample, inspired by diffusion model training~\cite{ho2020denoising, song2020denoising}.
Experiments in \autoref{tab:ablation_dataset_injection} imply that this strategy is still suboptimal, because it also cannot provide an appropriate hint level for different samples.   

\textbf{Adaptive hint level.}
To address this problem, we propose to adaptively adjust the hint level based on the difficulty level of different samples.
To this end, this strategy extends the original GRPO's single group output per sample~($G$ outputs per group) to $M$ groups, and assigns different hint ratios to each group.
Specifically, let $\{ \alpha_i \}_{i=1}^{M}$ denote the hint ratios for these $M$ groups, then $\alpha_i$ is set to $\frac{i - 1}{M}$.
As the hint ratio progressively increases, the sample difficulty steadily decreases from group $1$ through group $M$.
Finally, this strategy selects the first group with existing correct answers~(in the order from group $1$ to group $M$) for training.
By selecting the most appropriate hint level, this strategy both avoids the low-data utilization problem and mitigates the issue of excessive hints preventing the MLLM from reasoning, thereby achieving optimal results.

\textit{Efficiency analysis:} Although this strategy increases the group number to $M$, the MLLM still uses only one group for training.
Besides, we use vLLM~\cite{kwon2023efficient} to significantly speed up generation.
Therefore, compared to the original strategy, this strategy only increases the training time by 20.5\% when $M$ is 2, as shown in {\color{red} S2.1} of the appendix.

\subsection{Text-Bias Calibration}

This work reveals a phenomenon~(named text-bias) that the MLLM trained with GRPO tends to directly reason from text condition while ignoring image condition, as shown in \autoref{fig:intro_text_bias}.
Besides, \autoref{fig:acc_ratio}~(a) demonstrates that as GRPO training continues, the accuracy of MLLM (with the image condition removed) on the test set also improves.
We suspect that this phenomenon stems from that many query texts in current MLLM reasoning datasets can fully describe the questions, leading the MLLM to rely solely on text.
However, when the query text is insufficient to describe the entire question, this type of reasoning will result in errors and lower performance.

To mitigate this text-bias problem, this work proposes text-bias calibration, which can directly emphasize the image condition \textbf{in test-time}.
Inspired by CFG~(classifier-free guidance)~\cite{ho2022classifier} in image generation, text-bias calibration first uses the MLLM to predict token logits with and without image condition, and then calibrates the token logits according to their differences.
Specifically, let $q_{\rm img}$ and $q_{\rm text}$ denote the image condition and text condition, then $\hat{\pi}_\theta(o_t|q_{\rm img}) = \pi_\theta(o_t|q_{\rm img}, q_{\rm text}, o_{<t})$ and $\hat{\pi}_\theta(o_t) = \pi_\theta(o_t|q_{\rm text}, o_{<t})$ represent the token logit predicted with and without image condition~(note that $\hat{\pi}_\theta$ abbreviates the original $\pi_\theta$).
Finally, the calibrated token logit $\hat{\pi}_\theta^{\rm calibrated}(o_t|q_{\rm img})$ is calculated as below following CFG:

\vspace{-0.5em}
\begin{equation}
\label{equa:calibrated_logit}
    \hat{\pi}_\theta^{\rm calibrated}(o_t|q_{\rm img}) \! = \! \hat{\pi}_\theta(o_t|q_{\rm img}) + \gamma \cdot (\hat{\pi}_\theta(o_t|q_{\rm img}) - \hat{\pi}_\theta(o_t)),
\nonumber
\end{equation}

where $\gamma$ is a hyper-parameter controlling the intensity of image condition.
Intuitively, this text-bias calibration method makes the calibrated token logits move away from $\hat{\pi}_\theta(o_t)$ and closer to the real $\hat{\pi}_\theta(o_t|q_{\rm img})$, thereby alleviating the problem of ignoring image condition.
Furthermore, we provide a theoretical analysis on text-bias calibration in {\color{red} S1} of the appendix, for a more comprehensive understanding.

\begin{table*}
\small
\renewcommand\arraystretch{0.9}
\centering
\setlength{\tabcolsep}{1.4mm}{
\begin{tabular}{c|*{8}{c}}
  \toprule

\textbf{Method} & \textbf{\small Geo170K} & \makecell{\small \textbf{MathVista}\\[-1pt] \textbf{(Geometry)}} & \makecell{\small \textbf{MMStar}\\[-1pt] \textbf{(Geometry)}} & \makecell{\small \textbf{MathVerse}\\[-1pt] \textbf{(Geometry)}} & \makecell{\small \textbf{Math-Vision}\\[-1pt] \textbf{(Geometry)}} & \makecell{\small \textbf{MM-Math}\\[-1pt] \textbf{(Geometry)}} & \makecell{\small \textbf{WeMath}\\[-1pt] \textbf{(Geometry)}} & \textbf{\small Average} \\

\midrule
\multicolumn{9}{c}{\textbf{\textit{Qwen2-VL-7B}}} \\
\midrule

{\small Original} & 30.63 & 44.50 & 40.52 & 27.92 & 10.89 & 8.73 & 35.52 & 30.40 \\
{\small SFT} & 37.53 & 41.66 & 37.07 & 14.47 & 2.86 & 1.95 & 26.84 & 25.50 \\
{\small Mulberry} & 33.55 & 52.17 & 42.24 & 17.68 & 6.06 & 10.69 & 42.07 & 32.08 \\
{\small Open-R1-Multimodal} & 35.68 & 45.55 & 40.52 & 28.78 & 11.43 & 6.78 & 38.22 & 31.56 \\
{\small R1-V} & 38.72 & 47.26 & 41.38 & 28.12 & 12.51 & 8.83 & 41.44 & 33.19 \\
{\small GRPO} & 38.46 & 48.82 & 42.24 & 30.10 & 12.02 & 10.37 & 40.52 & 33.92 \\
{\small \bf Hint-GRPO} & 45.62 & 52.77 & 43.97 & 31.68 & 14.38 & 14.35 & 45.23 & 37.60 \\
{\small \bf Debiased Hint-GRPO} & \textbf{46.68} & \textbf{54.19} & \textbf{45.69} & \textbf{32.18} & \textbf{14.99} & \textbf{14.61} & \textbf{45.86} & \textbf{38.55} \\

\midrule
\multicolumn{9}{c}{\textbf{\textit{Qwen2.5-VL-3B}}} \\
\midrule

{\small Original} & 35.41 & 47.50 & 41.38 & 26.17 & 8.08 & 11.69 & 39.66 & 32.17 \\
{\small SFT} & 43.24 & 46.64 & 43.10 & 20.03 & 8.62 & 2.29 & 32.64 & 30.40 \\
{\small Open-R1-Multimodal} & 48.67 & 45.88 & 44.83 & 27.44 & 12.71 & 14.68 & 40.75 & 35.10 \\
{\small R1-V} & 47.48 & 48.67 & 47.41 & 31.42 & 14.58 & 12.10 & 42.07 & 36.55 \\
{\small GRPO} & 45.49 & 49.84 & 48.27 & 30.88 & 14.48 & 12.35 & 43.45 & 36.83 \\
{\small \bf Hint-GRPO} & 53.32 & 54.79 & 51.72 & 33.68 & 17.09 & 16.73 & 44.89 & 40.88 \\
{\small \bf Debiased Hint-GRPO} & \textbf{55.31} & \textbf{56.11} & \textbf{52.59} & \textbf{34.09} & \textbf{17.39} & \textbf{17.51} & \textbf{46.78} & \textbf{41.95} \\

\bottomrule
\end{tabular}}
\caption{Experiment results of two base MLLMs on the geometry reasoning tasks. Bold font denotes the best result.}
\vspace{-1em}
\label{tab:benchmark_geometry}
\end{table*}

\section{Experiments \label{sec:experiments}}

\noindent \textbf{Implementation details.}
Following existing GRPO methods~(\eg, R1-V~\cite{chen2025r1v}, Open-R1-Multimodal~\cite{open-r1-multimodal}), we conduct the main experiments on mathematical~(geometry) reasoning tasks, using Qwen2-VL-7B and Qwen2.5-VL-3B as base models.
Besides, we also follow LLaVA-o1~\cite{xu2024llava} to conduct experiments on the universal multimodal reasoning tasks, using Llama-3.2-11B-Vision as base model.
During training, we adopt AdamW optimizer with a learning rate of 5e-5, and train the model on 8 GPUs for 2 epochs~(following R1-V) with a batch size of 1 per GPU.
A system prompt is used to instruct the model to generate responses in the format of ``$<$think$>$\textit{\{Reasoning Steps}\}$<$/think$>$ $<$answer$>$\textit{\{Reasoning Result\}}$<$/answer$>$''.
We also use DeepSpeed~\cite{aminabadi2022deepspeed, rajbhandari2020zero} to facilitate the model training through ZeRO-3 optimization.
Besides, we use the vLLM package~\cite{kwon2023efficient} to accelerate the generation process in GRPO, allocating 1 GPU for generation and 7 GPUs for training.
The hyper-parameter $M$ for adaptive hint is set to 3, and the hyper-parameter $\gamma$ for text-bias calibration is set to 0.8.

\vspace{0.5em}
\noindent \textbf{Training dataset.}
For the geometry reasoning, we extract geometry samples from the LLaVA-CoT dataset~\cite{xu2024llava} for training, with a total size of 7840.
For the universal multimodal reasoning, we use the whole LLaVA-CoT dataset for training, with a total size of 100,000.

\vspace{0.5em}
\noindent \textbf{Test benchmark.}
For the geometry reasoning, we follow R1-V to use the same subset of Geo170K dataset~\cite{gao2023g} for evaluation.
Besides, we also incorporate other geometry data from existing datasets~(MathVista~\cite{lu2023mathvista}, MMStar~\cite{chen2024we}, MathVerse~\cite{zhang2024mathverse}, Math-Vision~\cite{wang2025measuring}, MM-Math~\cite{sun2024mm}, and WeMath~\cite{qiao2024we}) for evaluation, towards a comprehensive comparison.
For the universal multimodal reasoning, we follow LLaVA-o1 to evaluate the model on 6 benchmarks: MMStar~\cite{chen2024we}, MMBench~\cite{liu2024mmbench}, MMVet~\cite{yu2023mm}, MathVista~\cite{lu2023mathvista}, AI2D~\cite{hiippala2021ai2d}, and Hallusion~\cite{guan2024hallusionbench}.

\vspace{0.5em}
\noindent \textbf{Baseline methods.}
For the geometry reasoning, we compare our method with the SFT baseline, PRM methods~(Mulberry), and existing GRPO methods~(Open-R1-Multimodal, R1-V).
Specifically, Open-R1-Multimodal and R1-V are both GRPO methods but are trained on different datasets: GEOQA\_R1V and open-r1-8k-verified respectively. 
For the universal multimodal reasoning, we compare our method with the influential LLaVA-o1.

\begin{table*}
\small
\renewcommand\arraystretch{0.9}
\centering
\setlength{\tabcolsep}{1.5mm}{
\begin{tabular}{c|*{7}{c}}
  \toprule

\textbf{Method} & \textbf{\small MMStar} & \textbf{\small MMBench} & \textbf{\small MMVet} & \textbf{\small MathVista} & \textbf{\small AI2D} & \textbf{\small Hallusion} & \textbf{\small Average} \\
\midrule

{\small Original} & 49.8 & 65.8 & 57.6 & 48.6 & 77.3 & 40.3 & 56.6 \\
{\small LLaVA-o1} & 57.6 & 75.0 & 60.3 & 54.8 & 85.7 & 47.8 & 63.5 \\
{\small \bf Ours} & \textbf{60.7} & \textbf{75.8} & \textbf{64.2} & \textbf{56.8} & \textbf{86.6} & \textbf{50.7} & \textbf{65.8} \\

\bottomrule
\end{tabular}}
\caption{Experiment results of Llama-3.2-11B-Vision on the universal multimodal reasoning tasks. Bold font denotes the best result.}
\vspace{-1em}
\label{tab:benchmark_universal}
\end{table*}

\subsection{Comparison Analysis}

\noindent \textbf{Mathematical~(Geometry) reasoning.}
\autoref{tab:benchmark_geometry} shows the comparison results of different methods on two base MLLMs~(Qwen2-VL-7B \& Qwen2.5-VL-3B) across seven datasets.  
Several conclusions can be drawn from \autoref{tab:benchmark_geometry}:

(1) Supervised fine-tuning~(SFT) reduces the performance of original MLLMs on \textbf{out-of-domain data}, indicating that SFT merely memorizes knowledge mechanically, without learning patterns of reasoning that can generalize to new problems.

(2) PRM method~(Mulberry) outperforms SFT, demonstrating that PRM can enhance the intermediate reasoning steps of MLLM to some extent.

(3) GRPO methods~(Open-R1-Multimodal, R1-V, and GRPO on our dataset) demonstrate superior performance to both PRM and SFT methods, implying that GRPO enables the MLLM to independently learn how to think and reason for tackling new problems.

(4) As analyzed in \autoref{sec:method}, GRPO methods suffer from the low data utilization and text-bias problems, hindering the model training.
Our proposed text-debiased Hint-GRPO can alleviate these two problems, achieving significantly superior performance to existing GRPO methods.

\vspace{0.5em}
\noindent \textbf{Universal multimodal reasoning.}
\autoref{tab:benchmark_universal} presents the comparison results of different methods on Llama-3.2-11B-Vision across six datasets.
Both trained on the LLaVA-CoT dataset, our text-debiased Hint-GRPO \textbf{outperforms} the original LLaVA-o1 (trained with SFT).
Nevertheless, the improvement of the GRPO method on universal multimodal reasoning is not as significant as in geometry reasoning.
We discover that this issue may stem from the accuracy estimation in GRPO training, \eg, in samples requiring bounding box localization, the model's responses can hardly match the ground-truth answers perfectly, making such samples all incorrect and invalid for GRPO training.
Therefore, a more robust accuracy estimation method is required for universal multimodal reasoning.
For example, regarding the aforementioned samples, IoU values can be used to measure the accuracy of model responses. 

\begin{figure}[t]
\centering
    \includegraphics[width=1.0\linewidth]{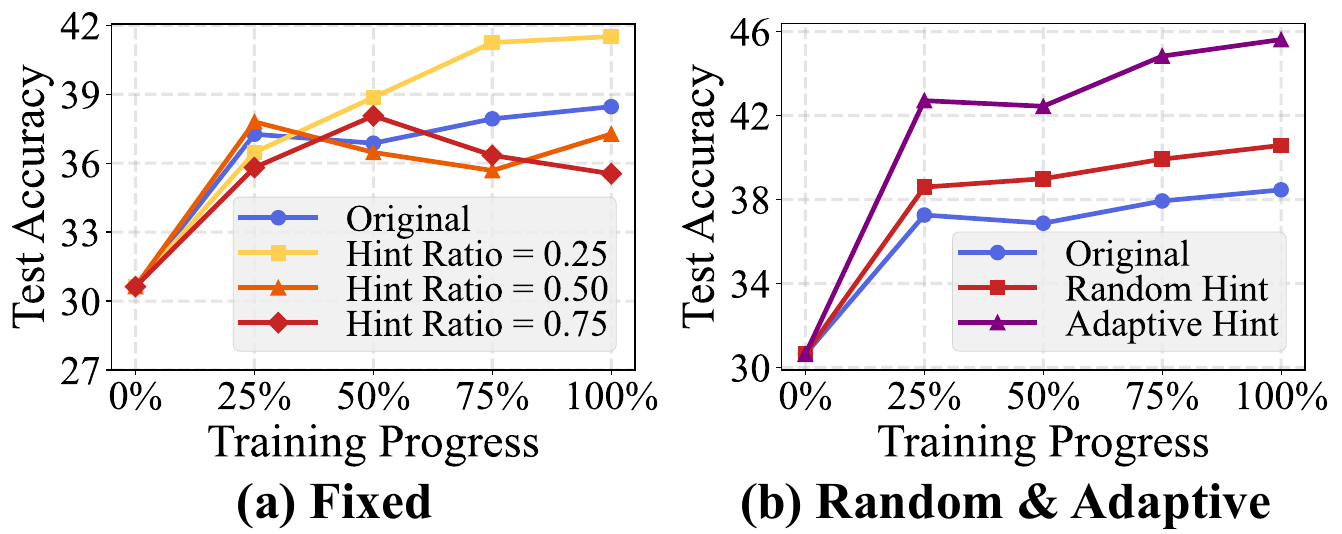}
\caption{Ablation experiments of hint adaptation strategy for Qwen2-VL-7B on the Geo170K dataset.}
\vspace{-1em}
\label{fig:ablation_hint_strategy}
\end{figure}

\subsection{Ablation Experiments}

\subsubsection{Hint-GRPO\label{sec:ablation_hint_grpo}}

This section provides the ablation experiments of three parts in Hint-GRPO:
(1) Dataset construction; (2) Hint injection method; (3) Hint adaptation strategy.

\vspace{0.5em}
\noindent \textbf{Dataset construction.}
We convert the multi-choice questions to fill-in-the-blank format in the original LLaVA-CoT dataset, and the original dataset and converted dataset are denoted as $\mathcal{D}_{\rm original}$ and $\mathcal{D}_{\rm new}$, respectively.
\autoref{tab:ablation_dataset_injection} shows that GRPO trained on $\mathcal{D}_{\rm new}$ outperforms GRPO trained on $\mathcal{D}_{\rm original}$ notably, by addressing the problem that the MLLM takes shortcuts~(\ie, random guessing) to correct answers rather than through reasoning.
Therefore, we use $\mathcal{D}_{\rm new}$ as the training dataset in the subsequent experiments.

\vspace{0.5em}
\noindent \textbf{Hint injection method.}
This work proposes two methods to inject the hint into the MLLM: \textit{hint injection in query}~(denoted as $\mathcal{I}_{\rm query}$) and \textit{hint injection in answer}~(denoted as $\mathcal{I}_{\rm answer}$).
\autoref{tab:ablation_dataset_injection} demonstrates that Hint-GRPO $\mathcal{I}_{\rm query}$ severely degrades performance, resulting from the inconsistency between query text \textbf{with hint} during training and query text \textbf{without hint} during testing.

\begin{table}
\small
\renewcommand\arraystretch{0.9}
\centering
\setlength{\tabcolsep}{0.4mm}{
\begin{tabular}{c|*{4}{c}}
  \toprule

\textbf{Method} & \textbf{\small MMStar} & \makecell{\small \textbf{MathVista}\\[-1pt] \textbf{(Geometry)}} & \makecell{\small \textbf{MMStar}\\[-1pt] \textbf{(Geometry)}} & \textbf{\small Avg.} \\
\midrule

{\small Original} & {\color{gray} 30.63} & {\color{gray} 44.50} & {\color{gray} 40.52} & {\color{gray} 40.04} \\
{\small GRPO + $\mathcal{D}_{\rm original}$} & 35.81 & 43.94 & 37.93 & 40.40 \\
{\small GRPO + $\mathcal{D}_{\rm new}$} & \textbf{38.46} & \textbf{48.82} & \textbf{42.24} & \textbf{44.59} \\
\midrule
{\small Hint-GRPO + $\mathcal{I}_{\rm query}$} & 41.64 & 47.17 & 39.66 & 43.91 \\
{\small Hint-GRPO + $\mathcal{I}_{\rm answer}$} & \textbf{45.62} & \textbf{52.77} & \textbf{43.97} & \textbf{48.78} \\

\bottomrule
\end{tabular}}
\caption{Ablation experiments of training dataset and hint injection method on Qwen2-VL-7B.}
\vspace{-1em}
\label{tab:ablation_dataset_injection}
\end{table}

\vspace{0.5em}
\noindent \textbf{Hint adaptation strategy.}
\autoref{fig:ablation_hint_strategy}~(a) shows the effect of hint ratio $\alpha$ on the strategy of \textbf{fixed hint level}, implying that a low-level hint ratio~(\eg, 0.25) can improve MLLM performance compared to the original GRPO, as hints enhance data utilization and facilitate the training.
However, excessive hint levels~(\eg, 0.50, 0.75) impair MLLM performance, which causes the MLLM to skip reasoning and undermine their reasoning capability.

\vspace{0.5em}
\autoref{fig:ablation_hint_strategy}~(b) compares the strategy of \textbf{random hint level} and \textbf{adaptive hint level}, indicating that random hint level increases MLLM performance compared to the original GRPO, due to its higher data utilization.
Nevertheless, compared to random hint level, adaptive hint level achieves more optimal performance, by selecting the most appropriate hint level for each sample according to its difficulty.   

Besides, \autoref{tab:ablation_M} shows the ablation experiments for the number of groups $M$, demonstrating that increasing $M$ can improve performance, but the performance saturates when $M$ becomes too large.
Therefore, based on efficiency considerations, $M$ is set to 3.

\begin{table}
\small
\renewcommand\arraystretch{0.9}
\centering
\setlength{\tabcolsep}{2.1mm}{
\begin{tabular}{c|*{5}{c}}
  \toprule

\textbf{Base Model} & \textbf{\small 0.0} & \textbf{\small 0.4} & \textbf{\small 0.8$^{\dagger}$} & \textbf{\small 1.2} & \textbf{\small 1.6} \\
\midrule

{\small Qwen2-VL-7B} & 45.62 & 46.29 & \textbf{46.68} & 46.02 & 44.69 \\
{\small Qwen2.5-VL-3B} & 53.32 & 54.77 & \textbf{55.31} & 54.38 & 53.58 \\

\bottomrule
\end{tabular}}
\caption{Ablation experiments of $\gamma$ for text-bias calibration on the Geo170K dataset. $\dagger$ denotes the selected one.}
\vspace{-1em}
\label{tab:ablation_gamma}
\end{table}
\begin{figure*}[t]
\centering
    \includegraphics[width=1.0\linewidth]{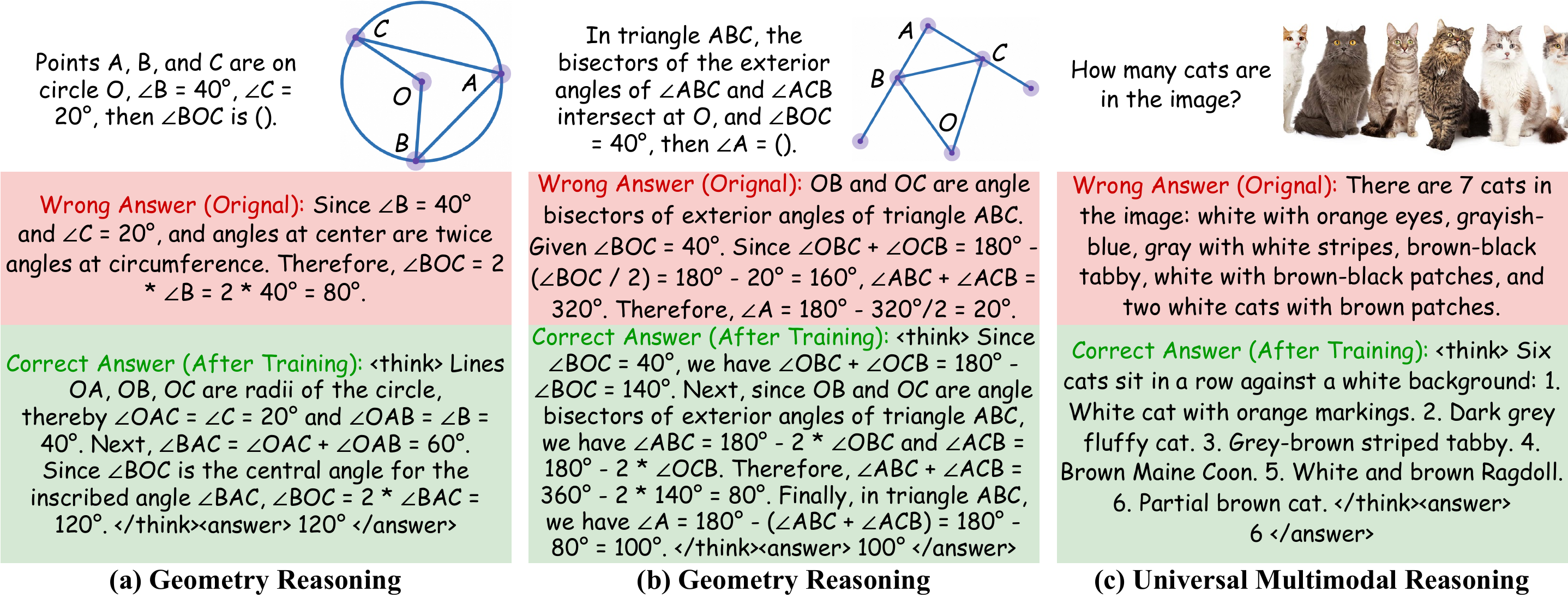}
\caption{Visualization examples of MLLMs' outputs before/after Hint-GRPO.}
\vspace{-1em}
\label{fig:vis_samples}
\end{figure*}
\begin{figure}[t]
\centering
    \includegraphics[width=1.0\linewidth]{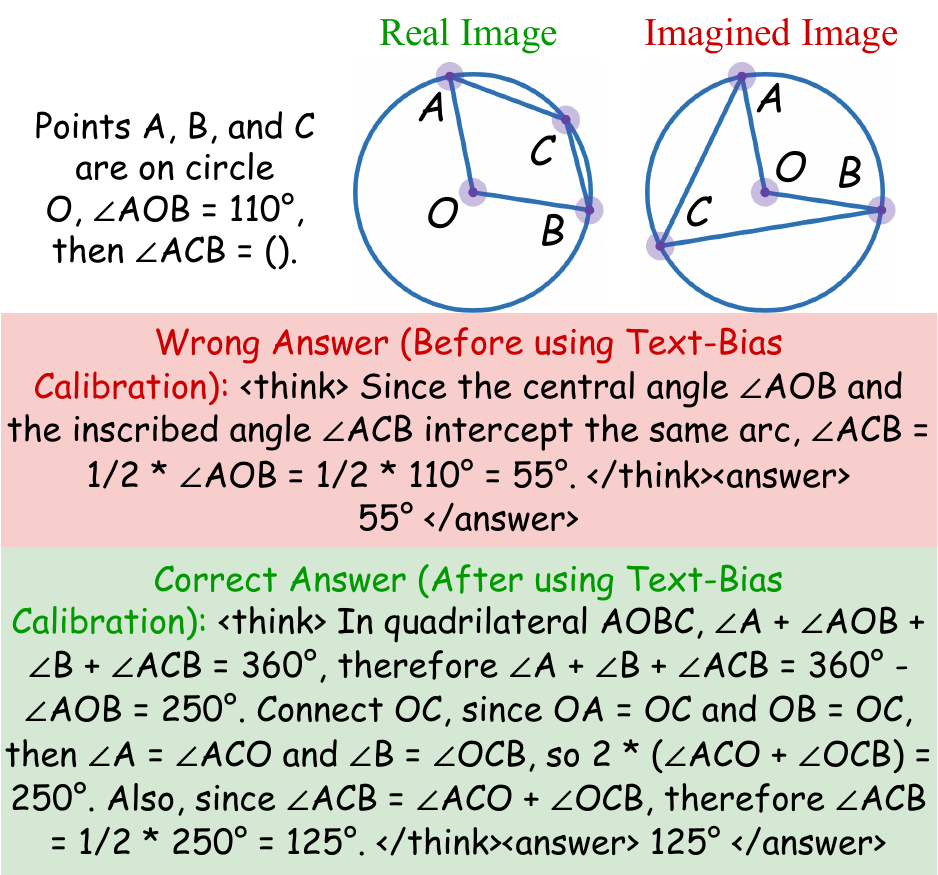}
\caption{Visualization example of Qwen2-VL-7B's outputs before/after text-bias calibration.}
\vspace{-1em}
\label{fig:vis_samples_text_bias}
\end{figure}

\subsubsection{Text-Bias Calibration}

As shown in \autoref{tab:benchmark_geometry}, text-debiased Hint-GRPO outperforms Hint-GRPO using the calibration operation in test-time, which increases the image conditioning intensity for alleviating the text-bias problem.
Besides, \autoref{tab:ablation_gamma} provides the ablation experiments for the effect of $\gamma$ on text-bias calibration, revealing that an excessively high $\gamma$ value degrades performance due to over-correction.

\subsection{Visualization Analysis}

\noindent \textbf{Visualization before/after Hint-GRPO.}
\autoref{fig:vis_samples} demonstrates the visualization examples of MLLMs' outputs before/after Hint-GRPO.
\autoref{fig:vis_samples}~(a) and (b) show that after Hint-GRPO training, the MLLM~(Qwen2-VL-7B) conducts more thorough analysis of difficult geometry problems, reaching correct solutions through extended reasoning processes~(within ``$<$think$>$'' and ``$<$/think$>$'' symbols).
Besides, \autoref{fig:vis_samples}~(c) presents that Hint-GRPO also achieves excellent performance in universal multimodal reasoning on Llama-3.2-11B-Vision.

\vspace{0.5em}
\noindent \textbf{Visualization before/after text-bias calibration.}
As shown in \autoref{fig:vis_samples_text_bias}, before text-bias calibration, the MLLM~(Qwen2-VL-7B) ignores the real image and uses the imagined image from text to generate the wrong answer.
After increasing the image conditioning intensity with text-bias calibration, the MLLM successfully concentrates on the real image and generates the correct answer from it.    

\begin{table}
\small
\renewcommand\arraystretch{0.9}
\centering
\setlength{\tabcolsep}{3.25mm}{
\begin{tabular}{c|*{4}{c}}
  \toprule

\textbf{Base Model} & \textbf{\small 1} & \textbf{\small 2} & \textbf{\small 3$^{\dagger}$} & \textbf{\small 4} \\
\midrule

{\small Qwen2-VL-7B} & 38.46 & 43.37 & 45.62 & \textbf{45.89} \\
{\small Qwen2.5-VL-3B} & 45.49 & 50.40 & \textbf{53.32} & 52.65 \\

\bottomrule
\end{tabular}}
\caption{Ablation experiments of $M$ for adaptive hint strategy on the Geo170K dataset. $\dagger$ denotes the selected one.}
\vspace{-1em}
\label{tab:ablation_M}
\end{table}

\section{Conclusion}

In this work, we identify and provide a thorough analysis of two problems that hinder the performance of GRPO on MLLM reasoning: Low data utilization and Text-bias.
In detail, low data utilization occurs when GRPO fails to obtain positive rewards for updating the MLLM on difficult samples.
Text-bias is a phenomenon where the MLLM disregards the image condition after GRPO training.
To tackle these two problems, we propose two key contributions: (1) Hint-GRPO, which improves data utilization by providing adaptive hints for samples of varying difficulty, and (2) text-bias calibration, which mitigates text-bias in test-time by calibrating the token prediction logits with image condition.
We conduct experiments on three base MLLMs across eleven datasets, showing that our proposed method achieves significantly superior performance to the original model, PRM methods, and existing GRPO methods.
We hope our method and dataset~(will be made publicly available) can contribute to the community of MLLM reasoning.

{
    \small
    \bibliographystyle{ieeenat_fullname}
    \bibliography{main}
}


\end{document}